\documentclass[10pt,twocolumn,letterpaper]{article}

\usepackage{compact}
\usepackage{times}
\usepackage{epsfig}
\usepackage{graphicx}
\usepackage{amsmath}
\usepackage{amssymb}

\usepackage[utf8]{inputenc} 
\usepackage[T1]{fontenc}    
\usepackage{url}            
\usepackage{booktabs}       
\usepackage{colortbl}       
\usepackage{amsfonts}       
\usepackage{microtype}      
\usepackage{placeins}
\usepackage{dsfont}
\usepackage{todonotes}
\usepackage[ruled,vlined]{algorithm2e}
\usepackage{xcolor}
\usepackage{placeins}
\usepackage{enumerate}
\usepackage[shortlabels]{enumitem}
\usepackage{caption}
\usepackage{subcaption}
\usepackage[export]{adjustbox}
\graphicspath{{./figures/}} 

\makeatletter
\newcommand{\settitle}{\@maketitle}
\makeatother

\usepackage{tikz}
\usetikzlibrary{arrows, arrows.meta, shapes, calc, positioning, decorations.pathreplacing, fit, backgrounds}

\usepackage[pagebackref=true,breaklinks=true,letterpaper=true,colorlinks,bookmarks=false]{hyperref}
\usepackage[capitalize]{cleveref}

\newcommand{\x}{\mathbf{x}}

\newcommand{\z}{\mathbf{z}}
\newcommand{\cond}{\mathbf{c}}
\newcommand{\uu}{\mathbf{u}}
\newcommand{\vv}{\mathbf{v}}
\newcommand{\bolda}{\mathbf{a}}
\newcommand{\boldb}{\mathbf{b}}
\newcommand{\boldL}{\mathbf{L}}

\DeclareMathSymbol{\sminus}{\mathbin}{AMSa}{"39}

\definecolor{vll-orange}{HTML}{E37238}
\definecolor{vll-green}{HTML}{96BF0D}
\definecolor{vll-dark}{HTML}{464646}
\definecolor{vll-light}{HTML}{757575}

\newcommand{\bftab}{\fontseries{b}\selectfont} 

\begin{document}

\title{Guided Image Generation with \\ Conditional Invertible Neural Networks}
\author{Lynton Ardizzone, Carsten Lüth, Jakob Kruse, Carsten Rother, Ullrich K\"othe \\
Visual Learning Lab Heidelberg}
\settitle

\begin{abstract}
    \noindent 
    In this work, we address the task of natural image generation guided by a conditioning input. We introduce a new architecture called conditional invertible neural network (cINN). 
    The cINN combines the purely generative INN model with an unconstrained feed-forward network, 
    which efficiently preprocesses the conditioning input into useful features.
    All parameters of the cINN are jointly optimized with a stable, maximum likelihood-based training procedure. 
    By construction, the cINN does not experience mode collapse and generates diverse samples, in contrast to e.g. cGANs.
    At the same time our model produces sharp images since no reconstruction loss is required, in contrast to e.g. VAEs.
    We demonstrate these properties for the tasks of MNIST digit generation and image colorization. 
    Furthermore, we take advantage of our bi-directional cINN architecture to explore and manipulate emergent properties of the latent space, such as changing the image style in an intuitive way.

    \noindent
    Code and appendix available at \\ \href{https://github.com/VLL-HD/FrEIA}{\color{black} \tt github.com/VLL-HD/FrEIA} \\
    Correspondence to \\ \texttt{\small lynton.ardizzone@iwr.uni-heidelberg.de}
\end{abstract}

\tikz[remember picture, overlay] \node [rotate=180, anchor=south west, font=\footnotesize, text=black!50] at ($(current page.south east)+(-2.22cm,2.1)$) {Quiz solution: Bottom row, center image};

\section{Introduction}
\begin{figure}[t!]
    \centering
    \includegraphics[width=\linewidth]{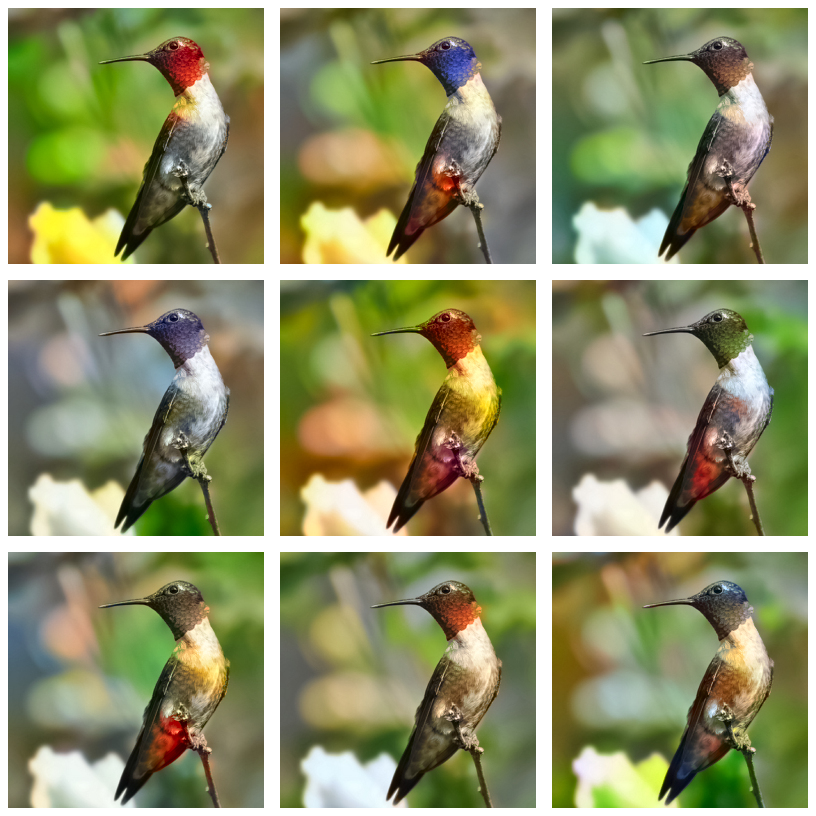}\vspace{-2mm}
	\caption{Diverse colorizations, which our network created for the same grayscale image. 
    One of them shows ground truth colors, but which?
    Solution at the bottom of the page.
	}%
	\vspace{-2mm}
	\label{fig:BIRD}
\end{figure}

\noindent 
Generative adversarial networks (GANs) produce ever larger and more realistic samples \cite{karras2017progressive, brock2018large}. Hence they have become the primary choice for a majority of image generation tasks.
As such, their conditional variants (cGANs) would appear to be the natural tool for conditional image generation as well, and they have successfully been applied in many scenarios \cite{ledig2017photo,miyato2018cgans}.
Despite numerous improvements, significant expertise and computational resources are required
to find a training configuration for large GANs that is stable, and produces diverse images.
A lack in diversity is especially common when the condition itself is an image, and special precautions have to be taken to avoid mode collapse.


Conditional variational autoencoders (cVAEs) do not suffer from the same problems.
Training is generally stable,
and since every data point is assigned a region in latent space, sampling yields the full variety of data seen during training.
However cVAEs come with drawbacks of their own:
The assumption of a Gaussian posterior on the decoder side implies an L2 reconstruction loss, which is known to cause blurriness. 
In addition, the partition of the latent space into diagonal Gaussians leads to either mode-mixing issues or regions of poor sample quality \cite{kingma2016improved}.
There has also been some success in combining aspects of both approaches for certain tasks, such as \cite{isola2017image, zhu2017toward, park2019semantic}.


We propose a third approach, by extending Invertible Neural Networks (INNs, \cite{dinh2016density,kingma2018glow,ardizzone2018analyzing}) for the task of {\em conditional} image generation, by adding conditioning inputs to their core building blocks.
INNs are neural networks which are by construction bijective, efficiently invertible, and have a tractable Jacobian determinant.
They represent transport maps between the input distribution $p(\x)$ and a prescribed, easy-to-sample-from latent distribution $p(\z)$.
During training, the likelihood of training samples from $p(\x)$ is maximized in latent space,
while at inference time, $\z$-samples can trivially be transformed back to the data domain.
Previously, INNs have been used successfully for unconditional image generation, e.g.~by \cite{dinh2016density} and \cite{kingma2018glow}. 

Unconditional INN training is related to that of VAEs, but it compensates for some key disadvantages:
Firstly, since reconstructions are perfect by design, no reconstruction loss is needed, and generated images do not become blurry.
Secondly, each $\x$ maps to exactly one $\z$ in latent space, and there is no need for posteriors $p(\z\,|\,\x)$.
This avoids the VAE problem of disjoint or overlapping regions in latent space.
In terms of training stability and sample diversity, INNs show the same strengths as autoencoder architectures, but with superior image quality.
We find that these positive aspects apply to conditional INNs (cINNs) as well.

One limitation of INNs is that their design restricts the use of some standard components of neural networks, such as pooling and batch normalization layers.
Our conditional architecture alleviates this problem, as the conditional inputs can be preprocessed by a 
{\it conditioning network} with a standard feed-forward architecture, which can be learned jointly with the cINN
to greatly improve its generative capabilities.
We demonstrate the qualities of cINNs for conditional image generation,
and uncover emergent properties of the latent space, for the tasks of conditional MNIST generation and diverse colorization of ImageNet.

Our work makes the following contributions:
\begin{itemize}
    \item We propose a new architecture called conditional invertible neural network (cINN), which combines an INN with an unconstrained feed-forward network for conditioning. It generates diverse images with high realism and thus overcomes limitations of existing approaches.

    \item We demonstrate a stable, maximum likelihood-based training procedure for jointly optimizing the parameters of the INN and the conditioning network.  
    
    \item We take advantage of our bidirectional cINN architecture to explore and manipulate emergent properties of the latent space. We illustrate this for MNIST digit generation and image colorization. 

\end{itemize}

\section{Related work}
\noindent
{\bf Conditional Generative Modeling.} 
Modern generative models learn to transform noise (usually sampled from multivariate Gaussians) into desired target distributions.
Methods differ by the model-family these transformations are picked from and by the losses determining optimal solutions.

Conditional generative adversarial networks (cGANs) \cite{mirza2014conditional} train a pair of neural networks: 
a {\em generator} transforms a pair of conditioning and noise vectors to images, and a {\em discriminator} penalizes unrealistic looking images.
The conditioning information is either concatenated to the noise \cite{mirza2014conditional}, or fed into the network via conditional batch-norm layers \cite{dumoulin2017learned,huang2017arbitrary,park2019semantic}. 
Ensuring diversity of the generated images (for fixed conditioning) appears to be challenging in this approach. 
Recent BigGANs \cite{brock2018large} successfully address this problem by using very large networks and batch sizes, but require parallel training on up to 512 TPUs.
PacGANs \cite{lin2018pacgan} employ augmented discriminators, which evaluate entire batches of real or generated images together rather than one image at a time.
CausalGANs \cite{kocaoglu2017causalgan} train two additional discriminator networks, called ``labeler'' and ``anti-labeler'', with the latter explicitly penalizing the lack of diversity.
Pix2pix \cite{isola2017image} addresses the important special case when the target is conditioned on an image in a different modality, e.g.~to generate satellite images from maps. 
In addition to the discriminator loss, it minimizes the L1 distance between  generated and ground-truth targets using a paired training set, which contains corresponding images from both modalities.
This leads to impressive image quality, but lack of diversity seems to be an especially hard problem in this case. 
In contrast, our method does not need explicit precautions to promote diversity.

Bidirectional architectures augment generator networks with complementary encoder networks that learn the generator's inverse and enable reconstruction losses, which exploit cycle consistency requirements.
Conditional variational autoencoders (cVAEs) \cite{sohn2015cvae} replace all distributions in a standard VAE \cite{kingma2013auto} by the appropriate conditional distributions, and are trained to minimize the evidence lower bound (ELBO loss).
Since variational distributions are typically Gaussian, the reconstruction penalty is equivalent to squared loss, resulting in rather blurry generated images.
This is avoided by AGE networks \cite{ulyanov2018it} and CycleGANs \cite{zhu2017cyclegan}, which combine standard cGAN discriminators with L1 reconstruction loss in the data domain, 
and bidirectional conditional GANs \cite{jaiswal2017bidirectional}, which extend the GAN discriminator to act on the distributions in data and latent space jointly.
SPADE \cite{park2019semantic}, building upon pix2pix and pix2pixHD \cite{wang2018high}, augments cGANs with additional VAE encoders to shape the latent space such that diversity is ensured.  

Instead of enforcing bijectivity through cycle losses,
invertible neural networks are bidirectional by design, since encoder and generator are realized by forward and backward processing within a single bijective model. 
We focus on architectures whose forward and backward pass require the same computational effort.
The coupling layer designs pioneered by NICE \cite{dinh2014nice} and RealNVP \cite{dinh2016density} emerged as very powerful and flexible model families under this restriction.
Using additive coupling layers, i-RevNets \cite{jacobsen2018irevnet} demonstrated that the lack of information reduction from data space to latent space does not cause overfitting.
The Glow architecture \cite{kingma2018glow} combines affine coupling layers with invertible 1x1 convolutions and achieves impressive attribute manipulations (e.g. age, hair color) in generated faces images.
This approach was recently generalized to video \cite{kumar2019videoflow}.

Thanks to tractable Jacobian determinants, the coupling layer architecture enables maximum likelihood training \cite{dinh2014nice,dinh2016density}, but experimental comparisons with other training methods are inconclusive so far.
For instance, \cite{danihelka2017comparison} found minimization of an adversarial loss to be superior to maximum likelihood training in RealNVPs, 
\cite{schirrmeister2018generative} trained i-RevNets in the same manner as adversarial auto-encoders, i.e.~with a discriminator acting in latent rather than data space,
and Flow-GANs \cite{grover2018flow} performed best using bidirectional training, a combination of maximum likelihood and adversarial loss.
On the other hand, maximum likelihood training worked well within Glow \cite{kingma2018glow}, and i-ResNets \cite{behrmann2018invertible} could even be trained with approximated Jacobian determinants.
In this work we reinforce the view that high-quality generative models can be trained by maximum likelihood loss alone.
To the best of our knowledge, we are the first to apply the coupling layer design for {\em conditional} generative models, 
with the exception of \cite{ardizzone2018analyzing}, who use it to compute posteriors for (relatively small) inverse problems, but do not consider image generation.

{\bf Colorization.}
State-of-the-art regression models for colorization produce visually near-perfect images \cite{iizuka2016let}, but do not account for the ambiguity inherent in this inverse problem.
To address this, models would ideally define a conditional distribution of plausible color images for a given grayscale input, instead of just returning a single ``best'' solution.

Popular existing approaches for diverse colorization predict per-pixel color histograms from a U-Net \cite{zhang2016colorful} or from hypercolumns of an adapted VGG network \cite{larsson2016learning}.
However, sampling from these local histograms independently can not lead to a spatially consistent colorization,
requiring additional heuristic post-processing steps to avoid artefacts.

%
In terms of generative models, both VAEs \cite{deshpande2017learning} and cGANs \cite{isola2017image, cao2017unsupervised} have been proposed for the task. 
However, their solutions do not reach the quality of the regression-based models, 
and cGANs in particular often lack diversity.
To compensate, modifications and extensions to generative approaches have been developed, 
such as auto-regressive models \cite{guadarrama2017pixcolor} and CRFs \cite{royer2017probabilistic}.
However, these methods are computationally very expensive and often unable to scale to realistic image sizes.

Conceptually closest to our proposed method is the work of \cite{ulyanov2018it}, 
where an encoder network maps color information to a latent space and a generator network learns the inverse transform,
both conditioned on the grayscale image.
Their experiments however are limited to a data set with only cars, and just three latent dimensions, leading to global, but no local diversity.

%
%
%
In contrast to the above, our flow-based cINN generates diverse colorizations in one standard feed-forward pass. It 
models the distribution of all pixels jointly, and allows for meaningful latent space manipulations.

\section{Method}
\noindent
Our method is an extension of the affine coupling block architecture established in \cite{dinh2016density}.
There, each network block splits its input $\uu$ into two parts $[\uu_1, \uu_2]$ and applies affine transformations between them that have strictly upper or lower triangular Jacobians:

\begin{equation}
\begin{aligned}
  \vv_1 &= \uu_1 \odot \exp\big(s_1(\uu_2)\big) + t_1(\uu_2) \\
  \vv_2 &= \uu_2 \odot \exp\big(s_2(\vv_1)\big) + t_2(\vv_1) \ .
  \label{eq:unconditional_forward}
\end{aligned}
\end{equation}
The outputs $[\vv_1, \vv_2]$ are concatenated again and passed to the next coupling block.
The internal functions $s_j$ and $t_j$ can be represented by arbitrary neural networks, and are only ever evaluated in the forward direction, even when the coupling block is inverted:
\begin{equation}
\begin{aligned}
  \uu_2 &= \big(\vv_2 - t_2(\vv_1) \big) \oslash \exp\big(s_2(\vv_1)\big)\\
  \uu_1 &= \big(\vv_1 - t_1(\uu_2) \big) \oslash \exp\big(s_1(\uu_2)\big) \ .
  \label{eq:unconditional_inverse}
\end{aligned}
\end{equation}
As shown in \cite{dinh2016density}, the logarithm of the Jacobian determinant for such a coupling block 
is simply the sum of $s_1$ and $s_2$ over image dimensions.

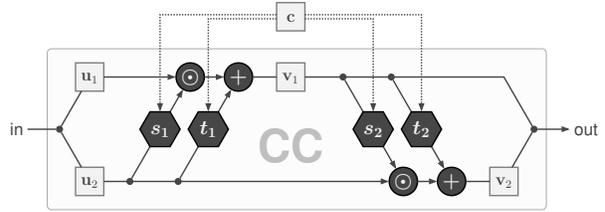
\begin{figure}[t]
	\centering
	\resizebox{\linewidth}{!}{ \begin{tikzpicture}[
    every node/.style = {inner sep = 0pt, outer sep = 0pt, anchor = center, align = center, font = \sffamily\large, text = black!75},
    var/.style = {rectangle, minimum width = 2em, minimum height = 2em, text depth = 0, line width = 1pt, draw = black!50, fill = black!5},
    op/.style = {circle, minimum width = 2em, text depth = 2pt, line width = 1pt, draw = black, fill = vll-dark, text = black!5, font = \Large\boldmath},
    nn/.style = {op, regular polygon, regular polygon sides = 6, minimum width = 1cm},
    dot/.style = {circle, minimum width = 5pt, fill = vll-dark},
    connect/.style = {line width = 1pt, draw = vll-dark},
    arrow/.style = {connect, -{Triangle[length=6pt, width=4pt]}}]

    \node [anchor = east] (in) at (0,0) {in};

    \node [dot] (split) at ([shift = {(in.east)}] 0:0.9) {};
    \draw [connect, shorten < = 3pt] (in) to (split);

    \node [var] (u1) at ([shift = {(split)}] 60:1.5) {$\uu_1$};
    \draw [connect] (split) to (u1);

    \node [var] (u2) at ([shift = {(split)}] -60:1.5) {$\uu_2$};
    \draw [connect] (split) to (u2);

    \node [dot] (d1) at ([shift = {(u2)}] 0:1) {};
    \node [dot] (d2) at ([shift = {(d1)}] 0:1.2) {};

    \node [op] (mult2) at ([shift = {(d1)}] 60:3) {$\odot$};
    \draw [arrow] (d1) to (mult2);
    \draw [arrow] (u1) to (mult2);

    \node [op] (add2) at ([shift = {(d2)}] 60:3) {$+$};
    \draw [arrow] (d2) to (add2);
    \draw [arrow] (mult2) to (add2);

    \node [var] (v1) at ([shift = {(add2)}] 0:1.3) {$\vv_1$};
    \draw [connect] (add2) to (v1);

    \node [dot] (d3) at ([shift = {(v1)}] 0:1.3) {};
    \node [dot] (d4) at ([shift = {(d3)}] 0:1.2) {};

    \node [op] (mult1) at ([shift = {(d3)}] -60:3) {$\odot$};
    \draw [arrow] (d3) to (mult1);
    \draw [arrow] (u2) to (mult1);

    \node [op] (add1) at ([shift = {(d4)}] -60:3) {$+$};
    \draw [arrow] (d4) to (add1);
    \draw [arrow] (mult1) to (add1);

    \node [var] (v2) at ([shift = {(add1)}] 0:1.3) {$\vv_2$};
    \draw [connect] (add1) to (v2);

    \node [dot] (cat) at ([shift = {(v2)}] 60:1.5) {};
    \draw [connect] (v2) to (cat);
    \path [connect] (v1) -- (v1.center -| v2.center) -- (cat);

    \node [anchor = west] (out) at ([shift = {(cat)}] 0:1) {out};
    \draw [arrow, shorten > = 3pt] (cat) to (out);

    \node [nn] (s2) at ([shift = {(d1)}] 60:1.5) {$s_1\vphantom{t}$};
    \node [nn] (t2) at ([shift = {(d2)}] 60:1.5) {$t_1$};
    \node [nn] (s1) at ([shift = {(d3)}] -60:1.5) {$s_2\vphantom{t}$};
    \node [nn] (t1) at ([shift = {(d4)}] -60:1.5) {$t_2$};

    \node [var] (cnet) at ([shift = {(v1)}] 90:1.5) {$\cond\vphantom{t}$};
    \draw [arrow, densely dotted] ([shift = {(cnet.west)}] 90:0.05) -- ([shift = {(cnet -| s2.north)}] 90:0.05) -- (s2.north);
    \draw [arrow, densely dotted] ([shift = {(cnet.west)}] -90:0.05) -- ([shift = {(cnet -| t2.north)}] -90:0.05) -- (t2.north);
    \draw [arrow, densely dotted] ([shift = {(cnet.east)}] -90:0.05) -- ([shift = {(cnet -| s1.north)}] -90:0.05) -- (s1.north);
    \draw [arrow, densely dotted] ([shift = {(cnet.east)}] 90:0.05) -- ([shift = {(cnet -| t1.north)}] 90:0.05) -- (t1.north);

    \coordinate (top left) at ($(u1.north west) + (-2em, 1em)$);
    \coordinate (bottom right) at ($(v2.south east) + (2em, -1em)$);
    \begin{scope}[on background layer]
        \node [rounded corners = 3pt, fill = black!1, draw = black!33, fit = (top left) (bottom right)] (bg) {};
        \node [font = \Huge \bfseries \sffamily, color = black!25, scale=1.3] (CC) at ([shift={(v1)}] -90:1.7) {CC};
    \end{scope}

\end{tikzpicture} }%
    \caption{One conditional affine coupling block (CC).}
    \label{fig:conditional-coupling}
\end{figure}

\subsection{Conditional invertible transformations}

\noindent
We adapt the design of \cref{eq:unconditional_forward,eq:unconditional_inverse} to produce a conditional version of the coupling block.
Because the subnetworks $s_j$ and $t_j$ are never inverted, we can concatenate conditioning data $\cond$ to their inputs without losing the invertibility, replacing $s_1(\uu_2)$ with $s_1(\uu_2, \cond)$ etc.
Our conditional coupling block design is illustrated in \cref{fig:conditional-coupling}.

In general, we will refer to a cINN with network parameters $\theta$ as $f(\x; \cond, \theta)$, and the inverse as $g(\z; \cond, \theta)$.
For any fixed condition $\cond$, the invertibility is given as 
\begin{equation}
  f^{-1}(\cdot \, ; \cond, \theta) = g(\cdot \, ; \cond, \theta).
\end{equation}

\subsection{Maximum likelihood training of cINNs}

\noindent
By prescribing a probability distribution $p_Z(\z)$ on latent space $Z$, the model $f$ assigns any input $\x$ a probability, 
dependent on both the network parameters $\theta$ and the conditioning $\cond$, through the change-of-variables formula:
\begin{equation}
  p_X(\x; \cond, \theta) = p_Z\left( f(\x; \cond, \theta) \right) 
  \left| \,\text{det}\!\left( \frac{\partial f}{\partial \x}\right)\right| \ .
  \label{eq:change_of_variables}
\end{equation}
Here, we use the Jacobian matrix ${\partial f}/{\partial \x}$. 
We will denote the determinant of the Jacobian, evaluated at some training sample $\x_i$, as
$ J_i \equiv \text{det}\big( {\partial f}/{\partial \x}|_{\x_i} \big)$.
Bayes' theorem gives us the posterior over model parameters as $p(\theta; \x, \cond) \propto p_X(\x; \cond, \theta) \cdot p_\theta(\theta)$.
Our goal is to find network parameters that maximize its logarithm, i.e.~we minimize the loss
\begin{equation}
  \mathcal{L} = \mathbb{E}_i\left[- \log\big(p_X(\x_i; \cond_i, \theta)\big)\right] - \log\big(p_\theta(\theta)\big),
\end{equation}
which is the same as in classical Bayesian model fitting.

Inserting \cref{eq:change_of_variables} with a standard normal distribution for $p_Z(\z)$, as well as a Gaussian prior on the weights $\theta$ with $1/2\sigma_\theta^2 \equiv \tau$, we obtain
\begin{equation}
  \mathcal{L} = \mathbb{E}_i\!\left[\frac{\| f(\x_i; \cond_i, \theta)\|_2^2}{2} - \log\big| J_i \big| \right] + \tau \| \theta \|_2^2 \ .
  \label{eq:ml_loss}
\end{equation}
The latter term represents L2 weight regularization, while the former is the {\it maximum likelihood loss}.

Training a network with this loss yields an estimate of the maximum likelihood network parameters $\hat\theta_\text{ML}$.
From there, we can perform conditional generation for a fixed $\cond$ by sampling $\z$ and using the inverted network $g$:
$  \x_\text{gen} = g(\z; \cond, \hat\theta_\text{ML})$, with $\z\sim p_Z(\z)$.

Training with the maximum likelihood method makes it virtually impossible for mode collapse to occur:
If any mode in the training set has low probability under the current guess $p_X(\x; \cond, \theta)$, 
the corresponding latent vectors will lie far outside the normal distribution $p_Z$ and receive big loss from the first L2-term in \cref{eq:ml_loss}.
In contrast, the discriminator of a GAN only supplies a weak signal, proportional to the mode's relative frequency in the training data,
so that the generator is not penalized much for ignoring a mode completely.

\subsection{Conditioning network}
\noindent
In complex settings, we expect that higher-level features of $\cond$ need to be extracted for the conditioning to be effective,
e.g.~global semantic information from an image as in \cref{sec:colorization}.
In such cases, feeding the condition $\cond$ directly into the cINN would place an unreasonable burden on the $s$ and $t$ networks,
as higher-level features would have to be re-learned in each coupling block.

To address this issue, we introduce an additional feed-forward {\it conditioning network} $h$,
which transforms the condition $\cond$ to some intermediate representation $\tilde{\cond} = h(\cond)$, and replace $\cond_i$ in \cref{eq:ml_loss} with $\tilde{\cond}_i = h(\cond_i)$.
The network $h$ can be pretrained, e.g.~by using features from a VGG architecture trained for image classification.
Alternatively or additionally, $h$ can be trained jointly with the cINN by propagating gradients from the maximum likelihood loss through the conditioning $\tilde{\cond}$. 
In this case, the conditioning network will learn to extract features which are particularly useful for embedding the cINN inputs $\x$ into latent variables $\z$.

\subsection{Important details}
\label{sec:architecture_details}
\noindent
For cINNs to match the performance of well-established architectures for conditional generation, 
we introduce a number of minor modifications and adjustments to the architecture and training procedure.
With these adaptions, our training setup is very stable and converges every time. 
Ablation results are presented in Sec.~\ref{sec:ablations}.

{\bf Noise as data augmentation.}
We add a small amount of noise to the inputs $\x$ as part of the standard data augmentation.
This helps to smooth out quantization artifacts in the input, and prevents sparse gradients 
when large parts of the image are completely flat (as e.g.~in MNIST).

{\bf Soft clamping of scale coefficients.}
We apply an additional nonlinear function to the scale coefficients $s$, of the form
\begin{equation}
  s_\text{clamp} = \frac{2 \alpha}{\pi} \text{arctan}\left(\frac{s}{\alpha}\right),
\end{equation}
which yields $s_\text{clamp} \approx s$ for $|s| \ll \alpha$ and $s_\text{clamp} \approx \pm \alpha$ for $|s| \gg \alpha$.
This prevents any instabilities stemming from exploding magnitude of the exponential $\exp(s_\text{clamp})$.
We find $\alpha = 1.9$ to be a good value for most architectures.

{\bf Initialization.}
Heuristically, we find that Xavier initialization \cite{glorot2010understanding} leads to stable training from the start. We experienced training instability when initial parameter values were too high.
Similar to \cite{kingma2018glow}, we also initialize the last convolution in all $s$ and $t$ subnetworks to zero, so training starts from an identity transform.

{\bf Soft channel permutations.}
We use random orthogonal matrices to mix the information between the channels.
This allows for more interaction between the two information streams $\uu_1, \uu_2$ in the coupling blocks. 
A similar technique was used in \cite{kingma2018glow}, but our matrices stay fixed throughout training and are guaranteed to be cheaply invertible.

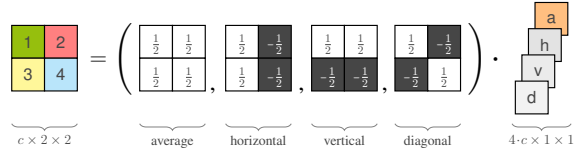
\begin{figure}
	\centering
	\resizebox{\linewidth}{!}{ \tikzstyle{box} = [rectangle, inner sep=0pt, outer sep=0pt, align=center, minimum width=1cm, minimum height=1cm, text depth=0, line width=1pt, draw=black, fill=white, font=\sffamily\Large, text=vll-dark]
\tikzstyle{dark} = [fill=vll-dark, text=white]
\tikzstyle{brace} = [decoration={brace, mirror, raise=0mm, amplitude=2mm}, decorate, black!50]
\tikzstyle{bracelabel} = [below=1mm, text=vll-dark, font=\large]

{\huge
\begin{math}
\begin{tikzpicture}[baseline=-0.65ex]
    \node [box, fill=vll-green] (1) at (-0.5, 0.5) {1};
    \node [box, fill=red!50] (2) at ( 0.5, 0.5) {2};
    \node [box, fill=yellow!50] (3) at (-0.5,-0.5) {3};
    \node [box, fill=cyan!25] (4) at ( 0.5,-0.5) {4};
    \draw [brace] (-1,-2) -- node [bracelabel] {$c \times 2 \times 2$} (1,-2);
\end{tikzpicture}\,
=
\Biggl(\;
\begin{tikzpicture}[baseline=-0.65ex]
    \node [box] (1) at (-0.5, 0.5) {$\frac{1}{2}$};
    \node [box] (2) at ( 0.5, 0.5) {$\frac{1}{2}$};
    \node [box] (3) at (-0.5,-0.5) {$\frac{1}{2}$};
    \node [box] (4) at ( 0.5,-0.5) {$\frac{1}{2}$};
    \draw [brace] (-1,-2) -- node [bracelabel] {average\vphantom{l}} (1,-2);
\end{tikzpicture}
\tikz[baseline=-0.65ex] \node[inner sep=0] at (0,-1) {\,,\;};
\begin{tikzpicture}[baseline=-0.65ex]
    \node [box] (1) at (-0.5, 0.5) {$\frac{1}{2}$};
    \node [box, dark] (2) at ( 0.5, 0.5) {$\sminus\frac{1}{2}$};
    \node [box] (3) at (-0.5,-0.5) {$\frac{1}{2}$};
    \node [box, dark] (4) at ( 0.5,-0.5) {$\sminus\frac{1}{2}$};
    \draw [brace] (-1,-2) -- node [bracelabel] {horizontal} (1,-2);
\end{tikzpicture}
\tikz[baseline=-0.65ex] \node[inner sep=0] at (0,-1) {\,,\;};
\begin{tikzpicture}[baseline=-0.65ex]
    \node [box] (1) at (-0.5, 0.5) {$\frac{1}{2}$};
    \node [box] (2) at ( 0.5, 0.5) {$\frac{1}{2}$};
    \node [box, dark] (3) at (-0.5,-0.5) {$\sminus\frac{1}{2}$};
    \node [box, dark] (4) at ( 0.5,-0.5) {$\sminus\frac{1}{2}$};
    \draw [brace] (-1,-2) -- node [bracelabel] {vertical} (1,-2);
\end{tikzpicture}
\tikz[baseline=-0.65ex] \node[inner sep=0] at (0,-1) {\,,\;};
\begin{tikzpicture}[baseline=-0.65ex]
    \node [box] (1) at (-0.5, 0.5) {$\frac{1}{2}$};
    \node [box, dark] (2) at ( 0.5, 0.5) {$\sminus\frac{1}{2}$};
    \node [box, dark] (3) at (-0.5,-0.5) {$\sminus\frac{1}{2}$};
    \node [box] (4) at ( 0.5,-0.5) {$\frac{1}{2}$};
    \draw [brace] (-1,-2) -- node [bracelabel] {diagonal} (1,-2);
\end{tikzpicture}
\;\Biggr)
\boldsymbol{\cdot}\!
\begin{tikzpicture}[baseline=-0.65ex]
    \node [box, fill=orange!50] (1) at ( 0.3, 1.2) {a};
    \node [box, fill=black!11] (2) at ( 0.1, 0.4) {h};
    \node [box, fill=black!8] (3) at (-0.1,-0.4) {v};
    \node [box, fill=black!5] (4) at (-0.3,-1.2) {d};
    \draw [brace] (-0.9,-2) -- node [bracelabel] {$4 \!\cdot\! c \times 1 \times 1$} (0.9,-2);
\end{tikzpicture}
\end{math}
} }%
	\vspace{-2mm}
    \caption{Haar wavelet downsampling reduces spatial dimensions \& separates lower frequencies (a) from high (h,v,d).}
    \label{fig:haar_wavelets}
\end{figure}
{\bf Haar wavelet downsampling.}
All prior INN architectures use checkerboard patterns for reshaping to lower spatial resolutions.
We find it helpful to instead perform downsampling with Haar wavelets \cite{haar1910wavelet},
which essentially decompose images into an average pooling channel as well as vertical, horizontal and diagonal derivatives, see \cref{fig:haar_wavelets}.
The three derivative channels contain high resolution information which we can split off early,
transforming only the remaining information further in later stages of the cINN.
This also contributes to mixing the variables between layers, complementing the soft permutations.
Similarly, \cite{jacobsen2018excessive} uses a discrete cosine transform as a final transformation in their INN, to replace global average pooling.

\begin{figure*}[!t!]
    \centering
    \includegraphics[width=1.0\textwidth]{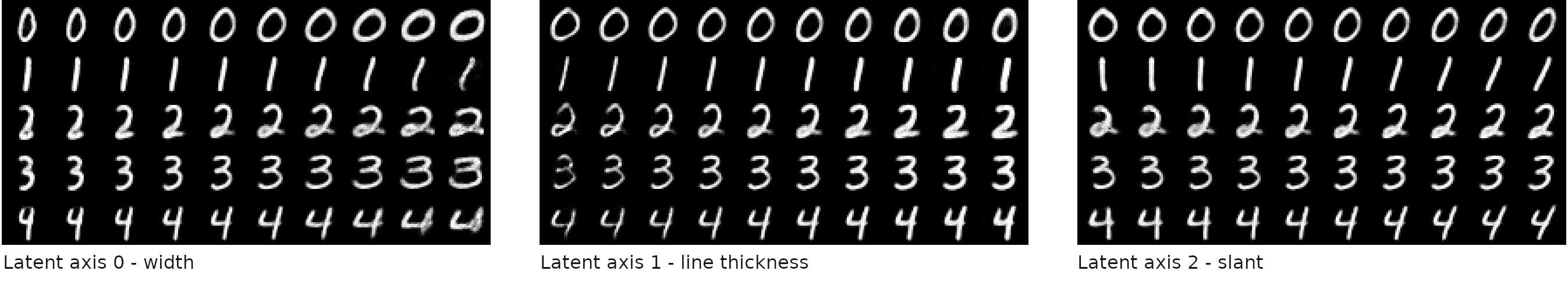}%
    \vspace{-4mm}%
	\caption{Axes in our MNIST model's latent space, which linearly encode the style attributes width, thickness and slant.}
	\label{fig:mnist_attributes}
\end{figure*}

\section{Experiments}
\noindent
We present results and explore the latent space of our models for two conditional image generation tasks: MNIST digit generation and image colorization. 

\subsection{Class-conditional generation for MNIST}

\begin{figure}[!b!]
	\centering
	\resizebox{0.9\linewidth}{!}{ \input{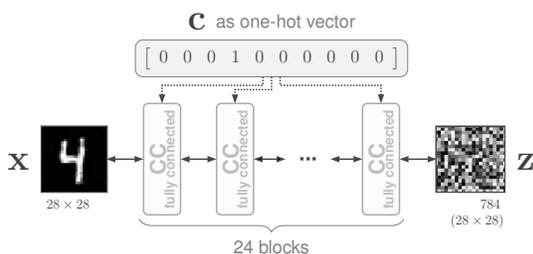} }%
	\vspace{-2mm}
    \caption{cINN model for conditional MNIST generation.}
    \label{fig:mnist-model}
\end{figure}

\noindent
As a first experiment, we perform simple class-conditional generation of MNIST digits.
We construct a cINN of 24 coupling blocks using fully connected subnetworks $s$ and $t$, 
which receive the conditioning directly as a one-hot vector (\cref{fig:mnist-model}).
No conditioning network $h$ is used.
For data augmentation we only add a small amount of noise to the images ($\sigma = 0.02$),
as described in \cref{sec:architecture_details}.

\begin{figure}
    \centering
	{ \def\svgwidth{\linewidth} 
\begingroup%
  \makeatletter%
  \providecommand\color[2][]{%
    \errmessage{(Inkscape) Color is used for the text in Inkscape, but the package 'color.sty' is not loaded}%
    \renewcommand\color[2][]{}%
  }%
  \providecommand\transparent[1]{%
    \errmessage{(Inkscape) Transparency is used (non-zero) for the text in Inkscape, but the package 'transparent.sty' is not loaded}%
    \renewcommand\transparent[1]{}%
  }%
  \providecommand\rotatebox[2]{#2}%
  \newcommand*\fsize{\dimexpr\f@size pt\relax}%
  \newcommand*\lineheight[1]{\fontsize{\fsize}{#1\fsize}\selectfont}%
  \ifx\svgwidth\undefined%
    \setlength{\unitlength}{248.57036681bp}%
    \ifx\svgscale\undefined%
      \relax%
    \else%
      \setlength{\unitlength}{\unitlength * \real{\svgscale}}%
    \fi%
  \else%
    \setlength{\unitlength}{\svgwidth}%
  \fi%
  \global\let\svgwidth\undefined%
  \global\let\svgscale\undefined%
  \makeatother%
  \begin{picture}(1,0.43843636)%
    \lineheight{1}%
    \setlength\tabcolsep{0pt}%
    \put(0,0){\includegraphics[width=\unitlength,page=1]{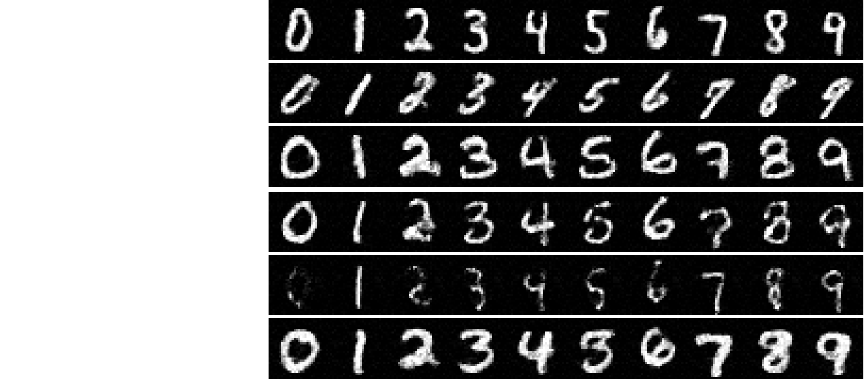}}%
    \put(-0.00048276,0.39422469){\color[rgb]{0,0,0}\makebox(0,0)[lt]{\lineheight{1.25}\smash{\begin{tabular}[t]{l}Tidy\end{tabular}}}}%
    \put(-0.00132122,0.32100269){\color[rgb]{0,0,0}\makebox(0,0)[lt]{\lineheight{1.25}\smash{\begin{tabular}[t]{l}Slanted, narrow\end{tabular}}}}%
    \put(-0.00132483,0.24653431){\color[rgb]{0,0,0}\makebox(0,0)[lt]{\lineheight{1.25}\smash{\begin{tabular}[t]{l}Slanted left, wide\end{tabular}}}}%
    \put(0.00019466,0.17073259){\color[rgb]{0,0,0}\makebox(0,0)[lt]{\lineheight{1.25}\smash{\begin{tabular}[t]{l}Messy\end{tabular}}}}%
    \put(-0.00246053,0.09862088){\color[rgb]{0,0,0}\makebox(0,0)[lt]{\lineheight{1.25}\smash{\begin{tabular}[t]{l}Faint\end{tabular}}}}%
    \put(-0.00246053,0.02270228){\color[rgb]{0,0,0}\makebox(0,0)[lt]{\lineheight{1.25}\smash{\begin{tabular}[t]{l}Bold\end{tabular}}}}%
  \end{picture}%
\endgroup%
 }
	\vspace{-6mm}
	\caption{MNIST samples from our cINN conditioned on digit labels.
	All ten digits within one row $(0, \dotsc, 9)$ were generated using the same latent code $\z$, but changing condition $\cond$.
	We see that each $\z$ encodes a single style consistently across digits, while varying $\z$ between rows leads to strong differences in writing style.
	}
	\label{fig:mnist_samples}
\end{figure}
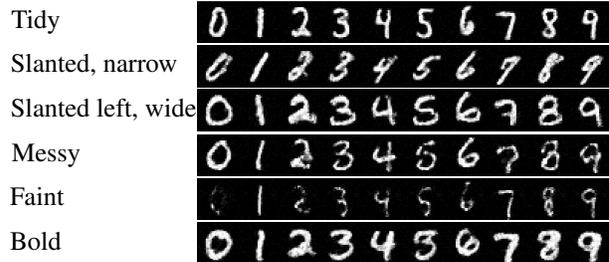

\begin{figure}
    \begin{flushright}
    \includegraphics[width=0.9\linewidth]{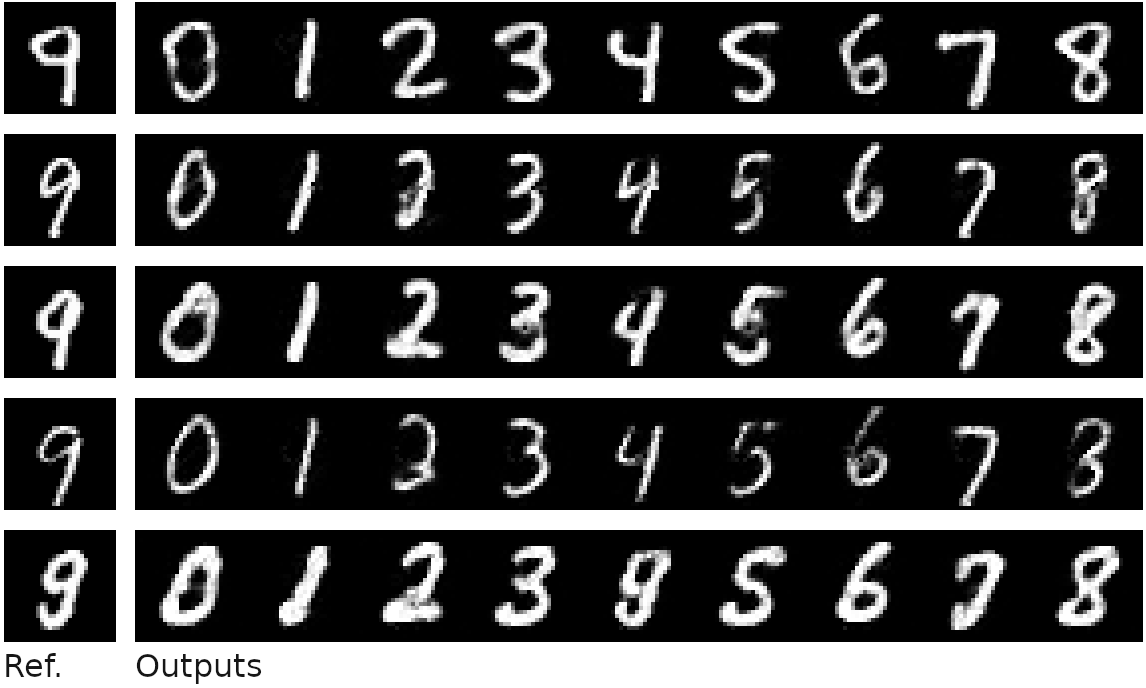}%
    \end{flushright}
    \vspace{-2mm}%
	\caption{To perform style transfer, we determine the latent code $\z = f(\x; \cond, \theta)$ of a validation image \emph{(left)}, then use the inverse network $g = f^{-1}$ with different conditions $\hat{\cond}$ to generate the other digits in the same style, $\hat{\x} = g(\z; \hat{\cond}, \theta)$.}
	\label{fig:mnist_transfer}
\end{figure}

Samples generated by the model are shown in \cref{fig:mnist_samples}. 
We find that the cINN learns latent representations that are shared across conditions $\cond$.
Keeping the latent vector $\z$ fixed while varying $\cond$ produces different digits in the same style.
This property, in conjunction with our network's invertibility, can directly be used for style transfer, as demonstrated in \cref{fig:mnist_transfer}.
This outcome is not obvious -- the trained cINN could also decompose into 10 essentially separate subnetworks, one for each condition.
In this case, the latent space of each class would be structured differently, and inter-class transfer of latent vectors would be meaningless.
The structure of the latent space is further illustrated in \cref{fig:mnist_attributes}, where we identify three latent axes with interpretable meanings.
Note that while the latent space is learned without supervision, we found the axes in a semi-automatic fashion:
We perform PCA on the latent vectors of the test set, without the noise augmentation, and manually identify meaningful directions in the subspace of the first four principal components.

\begin{figure*}[!t!]
	\centering
	\resizebox{\textwidth}{!}{ \input{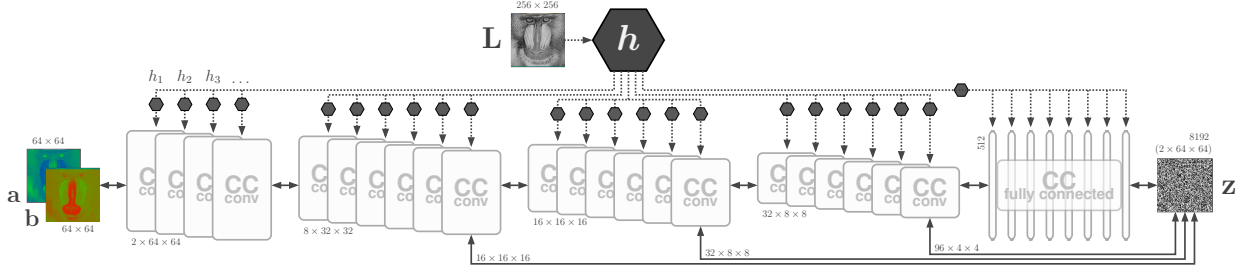} }%
    \caption{cINN model for diverse colorization.
    The conditioning network $h$ consists of a truncated VGG \cite{simonyan14vgg} pretrained to predict colors on ImageNet,
    with separate convolutional heads $h_1, h_2, h_3, \dots$ tailoring the extracted features to each individual conditional coupling block (CC).
    After each group of coupling blocks, we apply Haar wavelet downsampling (\cref{fig:haar_wavelets}) to reduce the spatial dimensions and,
    where indicated by arrows, split off parts of the latent code $\z$ early.
    }
    \label{fig:colorization-model}
\end{figure*}

\subsection{Diverse ImageNet colorization}
\label{sec:colorization}
\noindent
For a more challenging task, we turn to colorization of natural images.
The common approach for this task is to represent images in $Lab$ color space and generate color channels $\bolda, \boldb$ by a model conditioned on the luminance channel $\boldL$.

We train on the ImageNet dataset \cite{russakovsky15imagenet}, again adding low noise to the $\bolda, \boldb$ channels ($\sigma = 0.05$).
As the color channels do not require as much resolution as the luminance channel, 
we condition on $256\times256$ pixel grayscale images, but generate $64 \times 64$ pixel color information.
This is in accordance with the majority of existing colorization methods.

As with most generative INN architectures, we do not keep the resolution and channels fixed throughout the network, for the sake of computational cost.
Instead, we use 4 resolution stages, as illustrated in \cref{fig:colorization-model}. 
At each stage, the data is reshaped to a lower resolution and more channels, 
after which a fraction of the channels are split off as one part of the latent code.
As the high resolution stages have a smaller receptive field and less expressive power, 
the corresponding parts of the latent vector encode local structures and noise.
More global information is passed on to the lower resolution sections of the cINN.

\begin{figure*}
    \centering
    \includegraphics[width=0.42\linewidth]{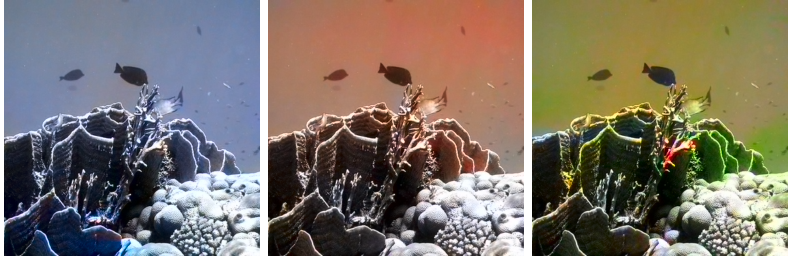}\hspace{5mm}%
    \includegraphics[width=0.42\linewidth]{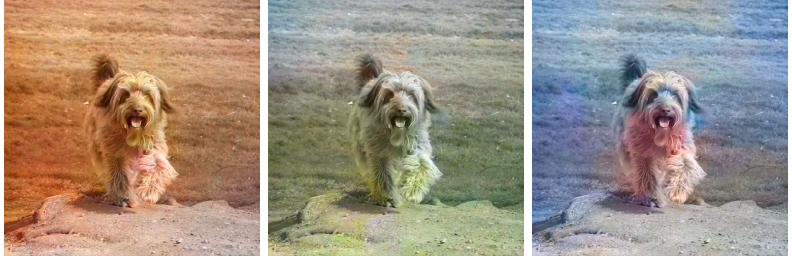}
    \vspace{-1mm}
	\caption{In an ablation study, we train a cINN using the grayscale image directly as conditional input, without a conditioning network $h$.
	The resulting colorizations largely ignore semantic content which leads to exaggerated diversity.
	More ablations are found in the appendix.
    }
	\label{fig:ablation}
\end{figure*}

For the conditioning network $h$, we start with the same VGG-like architecture and pretraining as \cite{zhang2016colorful},
i.e.~we pre-train the network to classify each pixel of the gray image into color bins.
By cutting off the network before the second-to-last convolution, we extract 256 feature maps of size $64\times64$ from the grayscale image $\boldL$.
We then add independent heads on top of this for each conditional coupling block in the cINN, indicated by small hexagons in \cref{fig:colorization-model}.
Thus each coupling block $k$ receives its own specialized conditioning $\tilde{\cond}^{(k)}_i = h_k\big(h(\cond_i)\big)$.
Each head consists of up to five strided convolutions, depending on its required output resolution, and a batch normalization layer.
The ablation study in \cref{fig:ablation} confirms that the conditioning network is 
necessary to capture semantic information.

We initially train the cINN and the $h_k$, keeping the parameters of the conditioning network $h$ fixed, for $30\,000$ iterations.
After this, we train both jointly until convergence, for 3 days on 3 Nvidia GTX1080 GPUs.
The Adam optimizer is essential for fast convergence,
and we lower the learning rate when the maximum likelihood loss levels off.

At inference time, we use joint bilateral upsampling \cite{kopf2007joint} to match the resolution of the generated color channels $\hat{\bolda}, \hat{\boldb}$
to that of the luminance channel $\boldL$.
This produces visually slightly more pleasing edges than bicubic upsampling, but has little to no impact on the results.
It was not used in the quantitative results table, to ensure an unbiased comparison.

The cINN compares favourably to existing methods, as shown in \cref{tab:results},
and has the best diversity and best-of-8 accuracy of the compared methods.
The cGAN apparently ignores the latent code, and relies only on the condition.
As a result, we do not measure any significant diversity,
in line with results from \cite{isola2017image}.

\begin{figure}
    \begin{flushright}
    \includegraphics[width=0.85\linewidth]{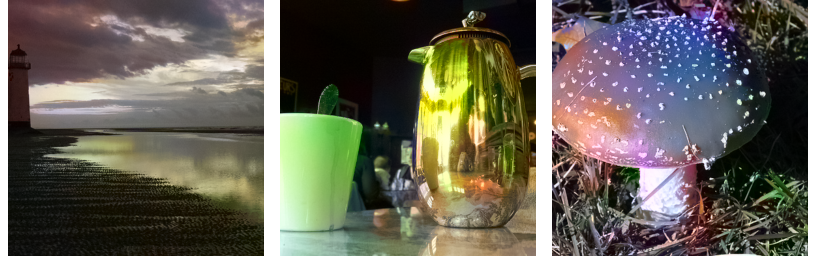}\\[1.0mm]
    \includegraphics[width=0.85\linewidth]{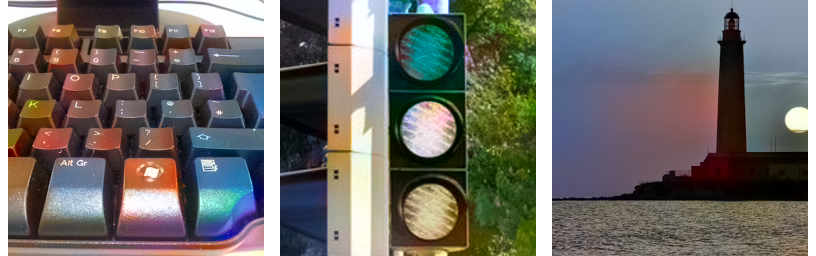} 
    \end{flushright}
	\caption{Failure cases of our method.
	{\em Top:} Sampling outliers.
	{\em Bottom:} cINN did not recognize an object's semantic class or the connectivity of occluded regions.
    \vspace{-1mm}
	}
	\label{fig:failure_cases}
\end{figure}

\begin{figure*}
    \centering
	{ \def\svgwidth{\textwidth} 
\begingroup%
  \makeatletter%
  \providecommand\color[2][]{%
    \errmessage{(Inkscape) Color is used for the text in Inkscape, but the package 'color.sty' is not loaded}%
    \renewcommand\color[2][]{}%
  }%
  \providecommand\transparent[1]{%
    \errmessage{(Inkscape) Transparency is used (non-zero) for the text in Inkscape, but the package 'transparent.sty' is not loaded}%
    \renewcommand\transparent[1]{}%
  }%
  \providecommand\rotatebox[2]{#2}%
  \newcommand*\fsize{\dimexpr\f@size pt\relax}%
  \newcommand*\lineheight[1]{\fontsize{\fsize}{#1\fsize}\selectfont}%
  \ifx\svgwidth\undefined%
    \setlength{\unitlength}{392.07207627bp}%
    \ifx\svgscale\undefined%
      \relax%
    \else%
      \setlength{\unitlength}{\unitlength * \real{\svgscale}}%
    \fi%
  \else%
    \setlength{\unitlength}{\svgwidth}%
  \fi%
  \global\let\svgwidth\undefined%
  \global\let\svgscale\undefined%
  \makeatother%
  \begin{picture}(1,0.18382352)%
    \lineheight{1}%
    \setlength\tabcolsep{0pt}%
    \put(0,0){\includegraphics[width=\unitlength,page=1]{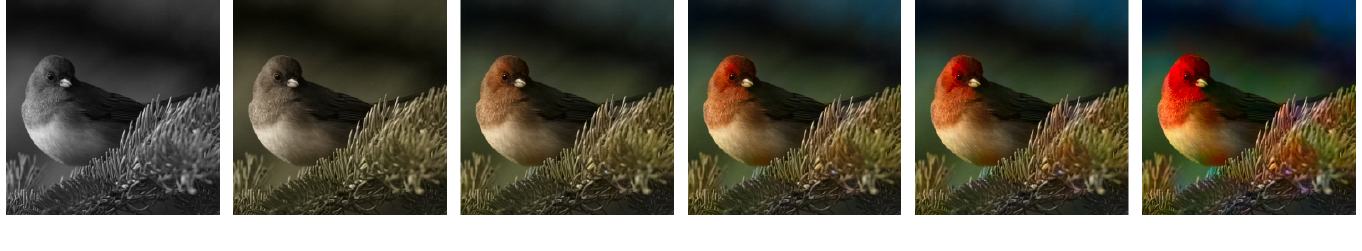}}%
    \put(0.00386467,0.00628917){\color[rgb]{0,0,0}\makebox(0,0)[lt]{\lineheight{1.25}\smash{\begin{tabular}[t]{l}Grayscale input\end{tabular}}}}%
    \put(0.17034886,0.00628017){\color[rgb]{0,0,0}\makebox(0,0)[lt]{\lineheight{1.25}\smash{\begin{tabular}[t]{l}$\z = 0.0 \cdot \z^*$\end{tabular}}}}%
    \put(0.337107,0.00628017){\color[rgb]{0,0,0}\makebox(0,0)[lt]{\lineheight{1.25}\smash{\begin{tabular}[t]{l}$\z = 0.7 \cdot \z^*$\end{tabular}}}}%
    \put(0.50386527,0.00628017){\color[rgb]{0,0,0}\makebox(0,0)[lt]{\lineheight{1.25}\smash{\begin{tabular}[t]{l}$\z = 0.9 \cdot \z^*$\end{tabular}}}}%
    \put(0.67062337,0.00628017){\color[rgb]{0,0,0}\makebox(0,0)[lt]{\lineheight{1.25}\smash{\begin{tabular}[t]{l}$\z = 1.0 \cdot \z^*$\end{tabular}}}}%
    \put(0.83738153,0.00628017){\color[rgb]{0,0,0}\makebox(0,0)[lt]{\lineheight{1.25}\smash{\begin{tabular}[t]{l}$\z = 1.25 \cdot \z^*$\end{tabular}}}}%
  \end{picture}%
\endgroup%
 }
    \includegraphics[width=\textwidth]{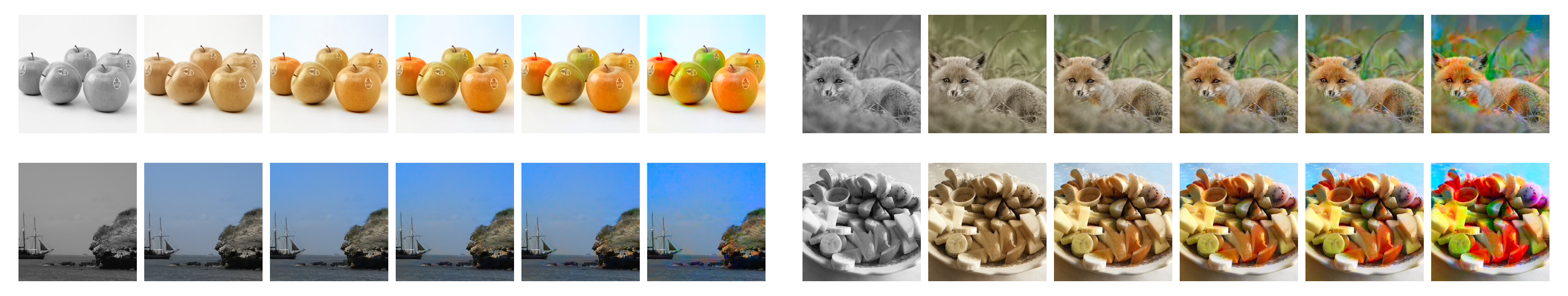}
	\caption{Effects of linearly scaling the latent code $\z$ while keeping the condition fixed.
	Vector $\z^*$ is ``typical'' in the sense that $\|\z^*\|^2=\mathbb{E}\big[\|\z\|^2\big]$, and results in natural colors.
	As we move closer to the center of the latent space ($\|\z\| < \|\z^*\|$), regions with ambiguous colors become desaturated, while less ambiguous regions (e.g.~sky, vegetation) revert to their prototypical colors.
	In the opposite direction ($\|\z\| > \|\z^*\|$), colors are enhanced to the point of oversaturation.
	}
	\label{fig:latent_temperature}
\end{figure*}
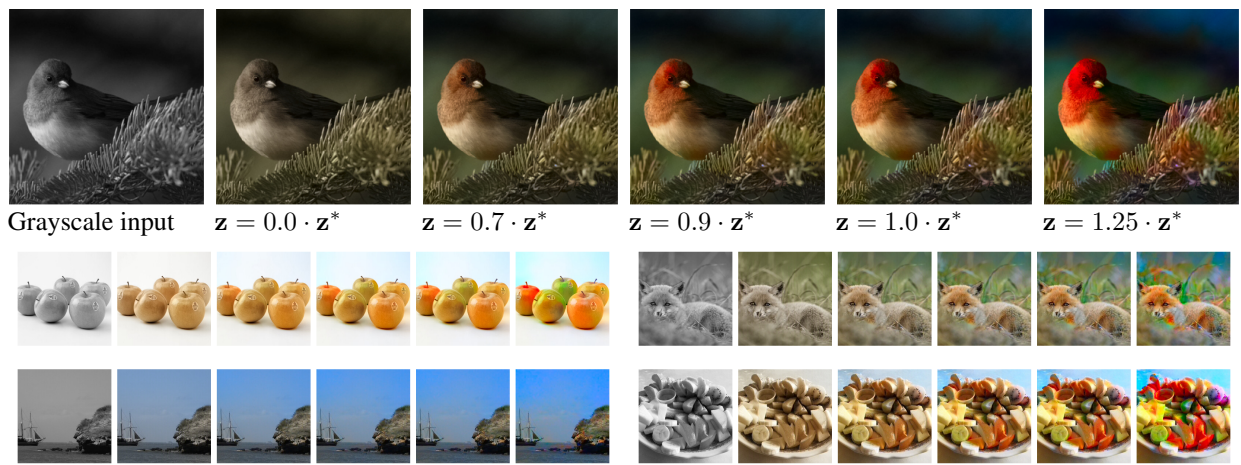

\begin{figure*}
    \centering
    \includegraphics[width=\textwidth]{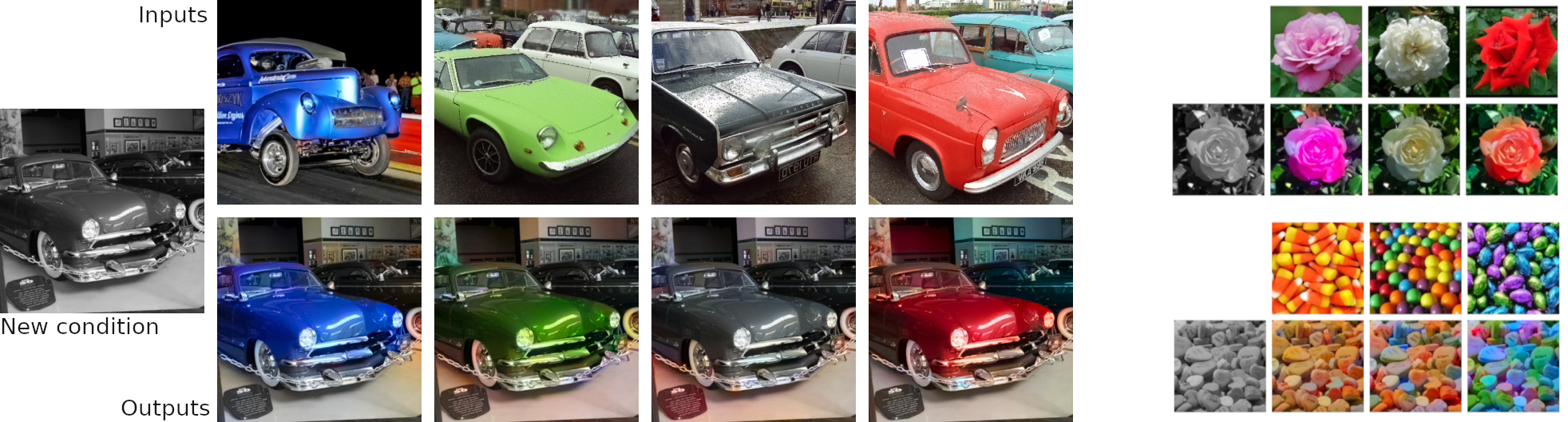}
	\caption{For color transfer, we first compute the latent vectors $\z$ for different color images $(\boldL,\bolda,\boldb)$ \emph{(top row)}. 
	We then send the same $\z$ vectors through the inverse network with a new grayscale condition $\boldL^*$ \emph{(far left)}
	to produce transferred colorizations $\bolda^*,\boldb^*$ \emph{(bottom row)}.
	Differences between reference and output color (e.g.~pink rose) can arise from mismatches between the reference colors $\bolda,\boldb$ and the intensity prescribed by the new condition $\boldL^*$.
	}
	\label{fig:latent_transfer}
\end{figure*}

In terms of FID score, the cGAN performs best, 
although its results do not appear more realistic to the human eye, cf.~\cref{fig:colorization_comparison}.
This may be due to the fact that FID is sensitive to outliers, which are unavoidable for a truly diverse method (see \cref{fig:failure_cases}), 
or because the discriminator loss implicitly optimizes for the similarity of deep CNN activations.
The VGG classification accuracy of generative methods is decreased compared to CNN, 
because occasional outliers may lead to misclassification.
Latent space interpolations and color transfer are shown in \cref{fig:latent_temperature,fig:latent_transfer}.

\subsection{Diverse bedrooms colorization}
To provide a simpler model for more in-depth experiments and ablations,
we additionally train a cINN for colorization on the LSUN bedrooms dataset \cite{yu2015lsun}.
We use a smaller model than for ImageNet, and train the conditioning network
jointly from scratch, without pretraining. 
Both the conditioning input, as well as the generated color channels have a resolution of $64\times 64$ pixels.
The entire model trains in under 4 hours on a single GTX 1080Ti GPU.

\begin{figure}
    \begin{center}
        \begin{tabular}{l | r | r}
            Metric & cINN & colorGAN \\
            \hline
            MSE best-of-8 &  \bftab 6.14  & 6.43  \\
            Variance      &  33.69 & \bftab 39.46 \\
            FID           & \bftab 26.48 & 28.31\\
        \end{tabular}
    \end{center}
    \vspace{2mm}

    \begin{tabular}{p{0.45\linewidth} p{0.5\linewidth}}
        cINN & COLORGAN
    \end{tabular}
    \includegraphics[width=\linewidth]{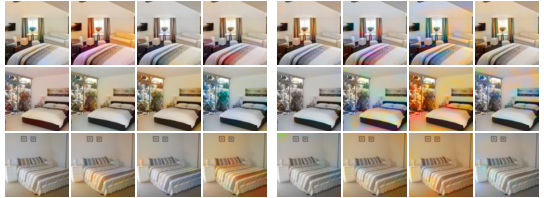}
    \caption{Quantitative and qualitative comparison between smaller cINN and colorGAN on LSUN bedrooms.
    The metrics used are explained in Table \ref{tab:results}.}
    \label{fig:colorgan}
\end{figure}

To our knowledge, the only diversity-enforcing cGAN architecture previously used 
for colorization is the colorGAN \cite{cao2017unsupervised}, 
which is also trained exclusively on the bedrooms dataset.
Training the colorGAN for comparison, we find 
it requires over 24 hours to converge stably, after multiple restarts.
The results are generally worse than those of the cINN,
as shown in Fig.~\ref{fig:colorgan}.
While the resulting pixel-wise color variance is slightly higher for the colorGAN, 
it is not clear whether this captures the true variance, 
or whether it is due to unrealistically colorful outputs, such as in the second row in Fig.~\ref{fig:colorgan}.

\subsection{Ablation of training improvements}
\label{sec:ablations}

\begin{figure}
    \centering
    \includegraphics[width=0.99\linewidth]{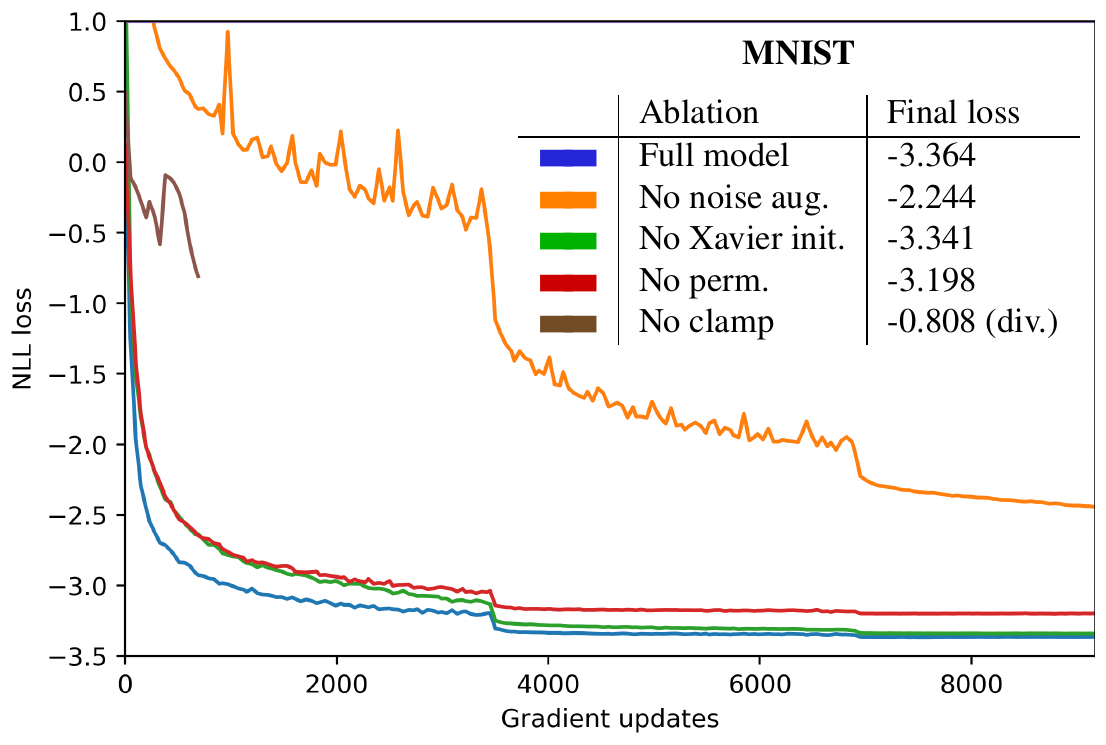}
    \includegraphics[width=0.99\linewidth]{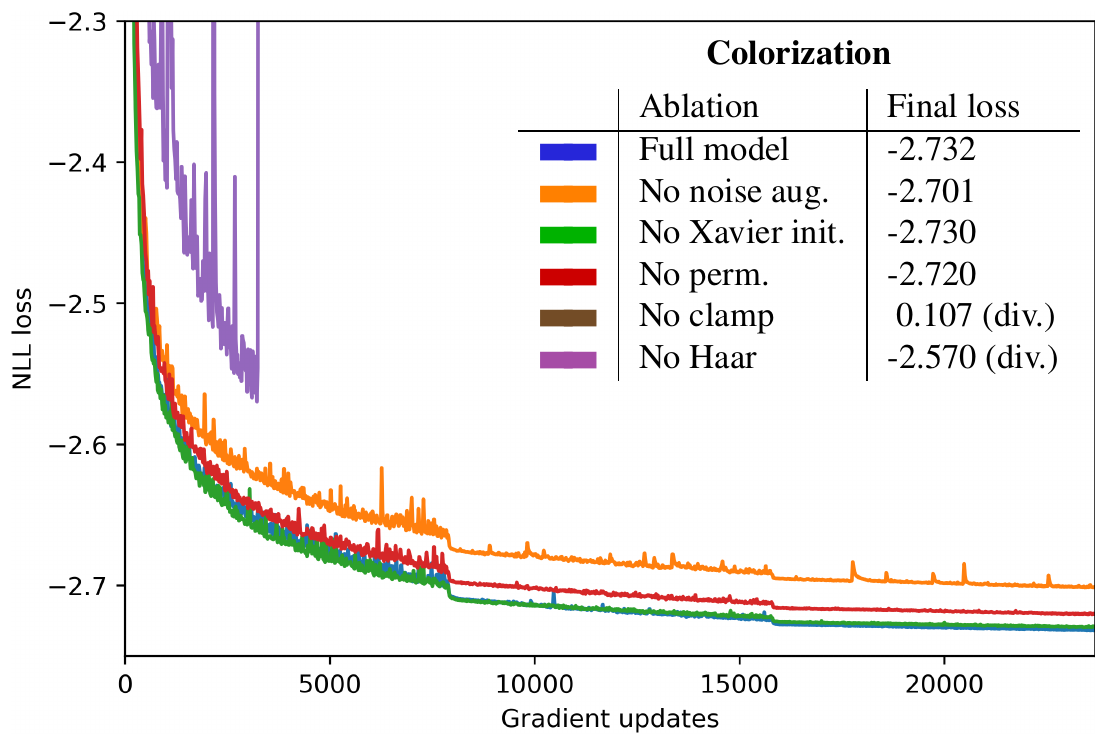}
	\caption{Training curves for each task, ablating the different improvements.}
	\label{fig:ablations}
\end{figure}

To demonstrate the improved stability and training speed through the improvements from Sec.~\ref{sec:architecture_details},
we perform ablations, see Fig.~\ref{fig:ablations}.
The ablations for colorization were performed for the LSUN bedrooms task, due to training speed.

We find that for stable training at Adam learning rates of $10^{-3}$, 
the clamping and Haar wavelet downsampling are strictly necessary.
Without these, the network has to be trained with much lower learning rates and more careful and specialized initialization, 
as used e.g. in \cite{kingma2018glow}.
Beyond this, the noise augmentation and permutations lead to the largest improvement in final result.
The effect of the noise is more pronounced for MNIST, as large parts of the image are completely black otherwise.
For natural images, dequantization of the data is likely to be the main advantage of the added noise.
The initialization only improves the final result by a small margin, but also converges noticeably faster.

\section{Conclusion and Outlook}
\noindent We have proposed a conditional invertible neural 
network architecture which enables 
guided generation of diverse images with high realism.
For image colorization, we believe that even better results 
can be achieved when employing latest tricks from large-scale GAN frameworks. 
Especially the non-invertible nature of the conditioning network 
make cINNs a suitable method for other computer vison tasks such 
as diverse semantic segmentation.

\section{Acknowledgments}
This work is supported by Deutsche Forschungsgemeinschaft (DFG) under
Germany's Excellence Strategy EXC-2181/1 - 390900948 (the Heidelberg STRUCTURES Excellence Cluster).
LA received funding by the Federal Ministry of Education and Research of Germany,
project `High Performance Deep Learning Framework' (No 01IH17002).
JK, CR and UK received financial support from the European Research Council (ERC) under the European Unions Horizon 2020 research and innovation program (grant agreement No 647769).
JK received funding by Informatics for Life funded by the Klaus Tschira Foundation.
Computations were performed on an HPC Cluster at the Center for
Information Services and High Performance Computing (ZIH) at TU Dresden.


\begin{table*}[t!]
\centering
\begin{tabular}{| l | r r r | r | r  r |}
\hline 
& cINN (ours)
& VAE-MDN \cite{deshpande2017learning} 
& cGAN \cite{isola2017image} 
& CNN \cite{iizuka2016let} 
& BW
& Ground truth \\
\hline 
MSE best of 8  
& {\bftab 3.53$\pm$0.04} 
&4.06$\pm$0.04  
& 9.75$\pm$0.06 
& 6.77 $\pm$0.05 
&  -- 
&  --  \\
Variance        
& {\bftab 35.2$\pm$0.3} 
& 21.1$\pm$0.2   
& 0.0$\pm$0.0 
&   --          
&   --          
&   --      \\
FID  \cite{heusel2017gans} 
& 25.13$\pm$0.30              
& 25.98$\pm$0.28         
& {\bftab 24.41$\pm$0.27}
& 24.95$\pm$0.27        
& $30.91\pm0.27$
&  14.69 $\pm$ 0.18 \\
VGG top 5 acc. 
& 85.00$\pm$0.48  
& 85.00$\pm$0.48 
& 84.62$\pm$0.53  
& {\bftab 86.86$\pm$0.41} 
& 86.02$\pm$0.43
& 91.66 $\pm$ 0.43\\
\hline 
\end{tabular}
\caption{Comparison of conditional generative models for diverse colorization.
We additionally compare to a state-of-the-art regression model (`CNN', no diversity),
and the grayscale images alone (`BW').
For each of 5k ImageNet validation images,
we compare the best pixel-wise MSE of 8 generated colorization samples, 
the pixel-wise variance between the 8 samples as an approximation of the diversity,
the Fréchet Inception Distance \cite{heusel2017gans} as a measure of realism,
and the top 5 accuracy of ImageNet classification performed on the colorized images, 
to check if semantic content is preserved by the colorization. 
}
\label{tab:results}
\end{table*}

\begin{figure}[b]
    \centering
    \includegraphics[width=0.95\linewidth]{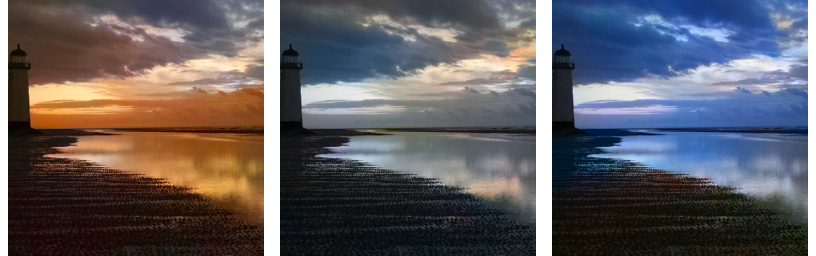} \\[1.2mm]
    \includegraphics[width=0.95\linewidth]{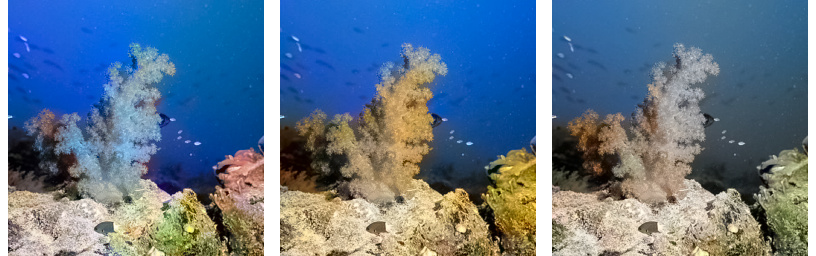} \\[1.2mm]
    \includegraphics[width=0.95\linewidth]{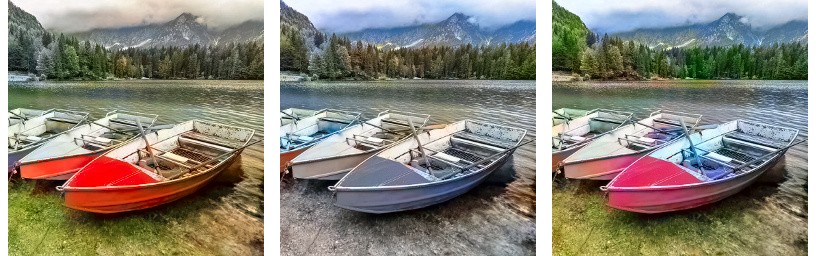} \\[1.2mm]
    \includegraphics[width=0.95\linewidth]{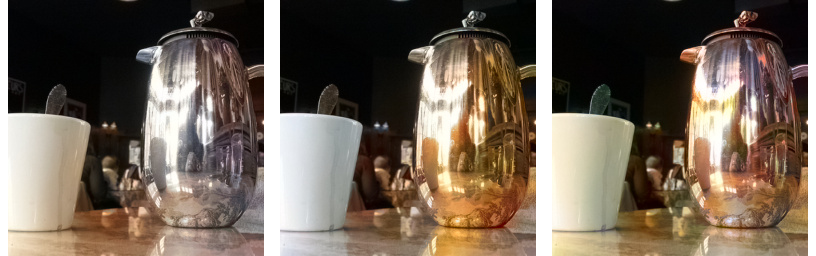} \\[1.2mm]
    \includegraphics[width=0.95\linewidth]{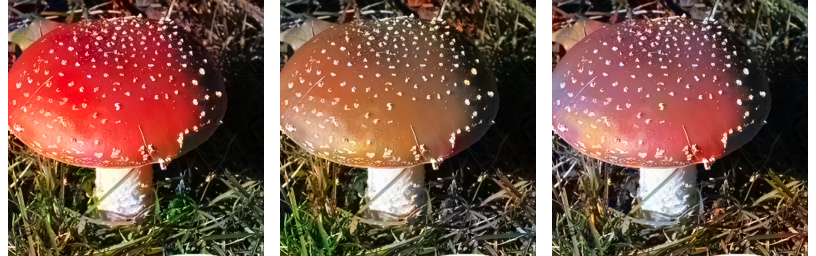} \\[1.2mm]
    \includegraphics[width=0.95\linewidth]{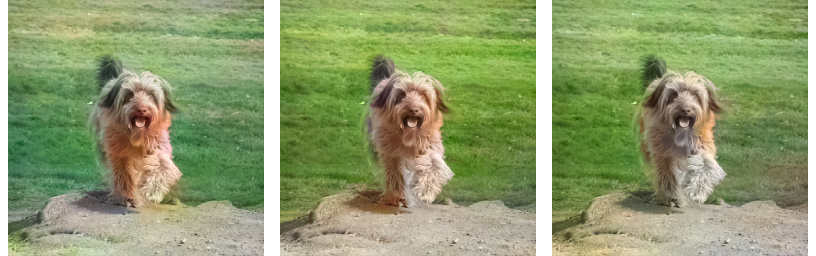}
    \vspace{-1mm}
	\caption{Diverse colorizations produced by our cINN.}
	\label{fig:colorization_examples}
\end{figure}

\begin{figure}
    \begin{flushright}
	{ \def\svgwidth{0.9\linewidth} 
\begingroup%
  \makeatletter%
  \providecommand\color[2][]{%
    \errmessage{(Inkscape) Color is used for the text in Inkscape, but the package 'color.sty' is not loaded}%
    \renewcommand\color[2][]{}%
  }%
  \providecommand\transparent[1]{%
    \errmessage{(Inkscape) Transparency is used (non-zero) for the text in Inkscape, but the package 'transparent.sty' is not loaded}%
    \renewcommand\transparent[1]{}%
  }%
  \providecommand\rotatebox[2]{#2}%
  \newcommand*\fsize{\dimexpr\f@size pt\relax}%
  \newcommand*\lineheight[1]{\fontsize{\fsize}{#1\fsize}\selectfont}%
  \ifx\svgwidth\undefined%
    \setlength{\unitlength}{405.38953886bp}%
    \ifx\svgscale\undefined%
      \relax%
    \else%
      \setlength{\unitlength}{\unitlength * \real{\svgscale}}%
    \fi%
  \else%
    \setlength{\unitlength}{\svgwidth}%
  \fi%
  \global\let\svgwidth\undefined%
  \global\let\svgscale\undefined%
  \makeatother%
  \begin{picture}(1,1.23500303)%
    \lineheight{1}%
    \setlength\tabcolsep{0pt}%
    \put(0,0){\includegraphics[width=\unitlength,page=1]{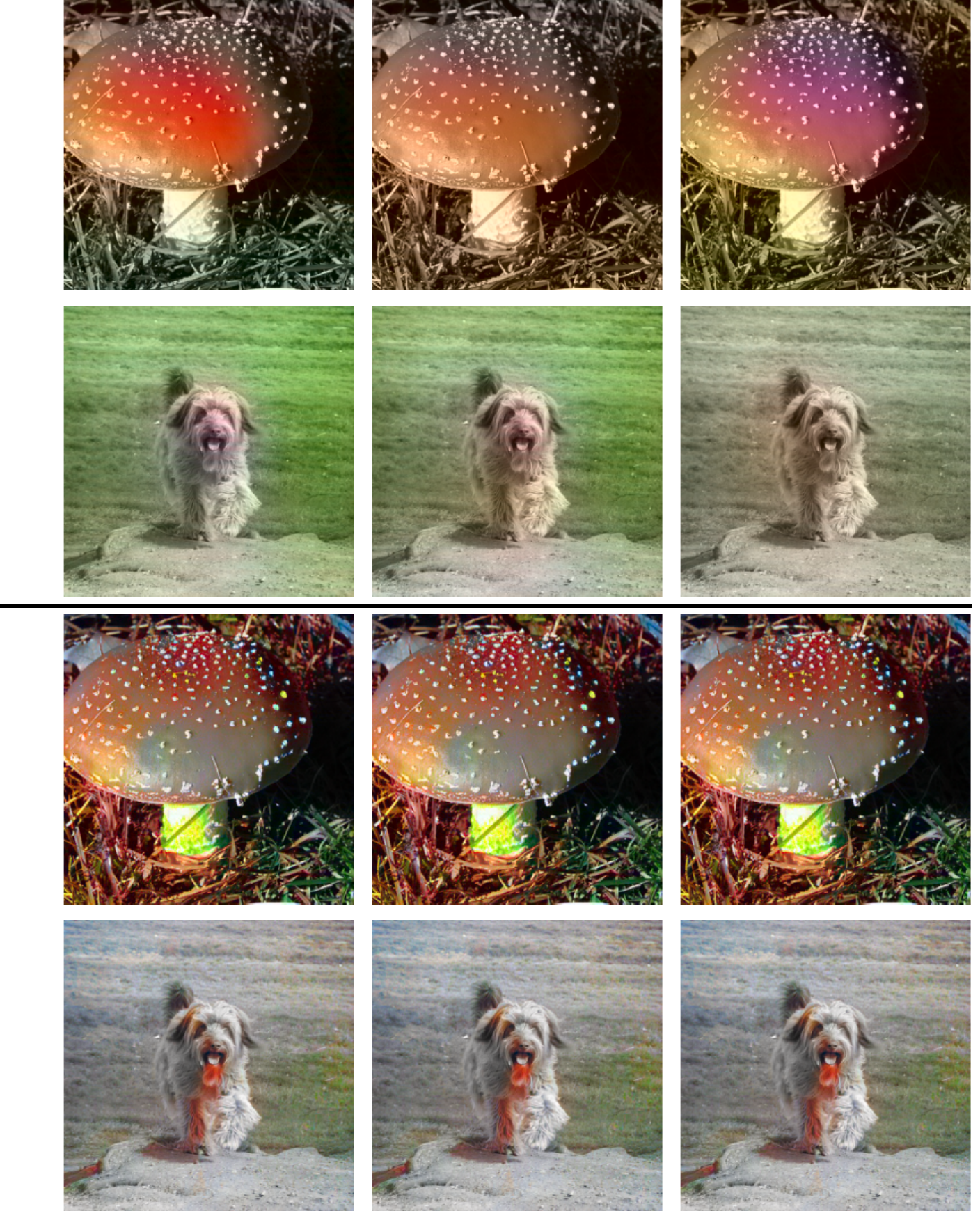}}%
    \put(0.05840434,0.89083765){\color[rgb]{0,0,0}\rotatebox{90}{\makebox(0,0)[lt]{\lineheight{1.25}\smash{\begin{tabular}[t]{l}VAE\end{tabular}}}}}%
    \put(0.05840434,0.24035675){\color[rgb]{0,0,0}\rotatebox{90}{\makebox(0,0)[lt]{\lineheight{1.25}\smash{\begin{tabular}[t]{l}cGAN\end{tabular}}}}}%
  \end{picture}%
\endgroup%
}
    \end{flushright}
	\caption{
	Alternative methods have lower diversity and lower quality, 
    suffering from inconsistencies within objects, or color blurriness and bleeding
    (compare \cref{fig:colorization_examples}, bottom).
	}
	\label{fig:colorization_comparison}
\end{figure}

\FloatBarrier
\clearpage
{\small
\bibliographystyle{ieee}
\bibliography{bibliography}
}

\end{document}